# ICGM-FRAX: Iterative Cross Graph Matching for Hip Fracture Risk Assessment using Dual-energy X-ray Absorptiometry Images

*Authors*

Chen Zhao[1], Anjum Shaik[2], Joyce H. Keyak[3], Nancy E. Lane[4], Jeffrey D. Deng[5], Kuan-Jui Su[6], Qiuying Sha[7], Hui Shen[6], Hong-Wen Deng[6], Weihua Zhou[2,8*]

*Institutions*

[1.] Department of Computer Science, Kennesaw State University, 680 Arntson Dr, Marietta, GA 30060

[2.] Department of Applied Computing, Michigan Technological University, 1400 Townsend Dr, Houghton, MI, 49931

[3.] Department of Radiological Sciences, Department of Biomedical Engineering, and Department of Mechanical and Aerospace Engineering, University of California, Irvine, CA 92697

[4.] Department of Internal Medicine and Division of Rheumatology, UC Davis Health, Sacramento, CA 95817

[5.] Geisel School of Medicine, Dartmouth College, Hanover, NH 03755

[6.] Division of Biomedical Informatics and Genomics, Tulane Center of Biomedical Informatics and Genomics, Deming Department of Medicine, Tulane University, New Orleans, LA 70112

[7.] Department of Mathematical Sciences, Michigan Technological University, Houghton, MI 49931

[8.] Center for Biocomputing and Digital Health, Institute of Computing and Cybersystems, and Health Research Institute, Michigan Technological University, Houghton, MI 49931

*\* Corresponding authors:*

Weihua Zhou, Ph.D.

Department of Applied Computing, Michigan Technological University,

1400 Townsend Dr, Houghton, MI, 49931, USA

Tel: 906-487-2666

E-Mail: whzhou@mtu.edu

**Abstract**:

Hip fractures represent a major health concern, particularly among the elderly, often leading decreased mobility and increased mortality. Early and accurate detection of at risk individuals is crucial for effective intervention. In this study, we propose Iterative Cross Graph Matching for Hip Fracture Risk Assessment (ICGM-FRAX), a novel approach for predicting hip fractures using Dual-energy X-ray Absorptiometry (DXA) images. ICGM-FRAX involves iteratively comparing a test (subject) graph with multiple template graphs representing the characteristics of hip fracture subjects to assess the similarity and accurately to predict hip fracture risk. These graphs are obtained as follows. The DXA images are separated into multiple regions of interest (RoIs), such as the femoral head, shaft, and lesser trochanter. Radiomic features are then calculated for each RoI, with the central coordinates used as nodes in a graph. The connectivity between nodes is established according to the Euclidean distance between these coordinates. This process transforms each DXA image into a graph, where each node represents a RoI, and edges derived by the centroids of RoIs capture the spatial relationships between them. If the test graph closely matches a set of template graphs representing subjects with incident hip fractures, it is classified as indicating high hip fracture risk. We evaluated our method using 547 subjects from the UK Biobank dataset, and experimental results show that ICGM-FRAX achieved a sensitivity of 0.9869, demonstrating high accuracy in predicting hip fractures.

# 1. Introduction

Hip fractures represent a major public health challenge, particularly for the aging population [1]. The consequences of hip fracture are severe, leading to increased mortality, long-term disability, and substantial healthcare costs [2]. Most hip fractures occur in individuals with osteoporosis, a condition characterized by reduced bone mineral density (BMD) and compromised bone quality. As the global population ages, the burden of hip fracture is expected to rise, underscoring the need for accurate and reliable methods for hip fracture risk assessment.

Areal bone mineral density (aBMD), measured through dual-energy X-ray absorptiometry (DXA), remains the gold standard for assessing bone health and determining fracture risk. However, while aBMD is widely used in clinical practice, its predictive ability for hip fracture is limited. Approximately half of individuals who experience a hip fracture are classified as not at risk based on aBMD values alone [3]. This limitation arises because aBMD only considers the overall bone mass and does not account for the heterogeneity in bone structure, geometry, or quality, which are critical factors in hip fracture risk. Recent studies have turned to advanced techniques that extract additional information from DXA images. Among these, Trabecular Bone Score (TBS) has shown promise as a complementary tool to aBMD. TBS, which evaluates bone microarchitecture, has been found to provide a more comprehensive assessment of bone quality, particularly in predicting fractures in populations with osteoporosis [4].

The advent of artificial intelligence (AI) and machine learning (ML) techniques has further advanced the potential for accurate hip fracture risk assessment. AI models, trained on features extracted from DXA scans, have demonstrated diagnostic accuracy comparable to expert clinicians in detecting hip fractures and predicting postoperative outcomes [5]. These models capitalize on the wealth of data available in DXA images, enabling a more nuanced understanding of fracture risk. However, while these methods have demonstrated high potential, there is still a need for approaches that can more effectively analyze the complex relationships within DXA images and improve interpretability.

In this paper, we propose a model using graph matching for hip fracture risk assessment on DXA scans. The overall architecture is shown in Figure 1. By comparing the similarities between a graph generated from the DXA scan for a test subject (the test graph) and a set of graphs generated from template DXA images, a label is assigned to the test graph based on the label distribution in the template DXA images using majority voting. We conducted our experiments using a subset of DXA images from the UK Biobank dataset, and the experimental results indicate that the proposed method achieved a sensitivity of 0.9869, significantly outperforming existing methods. In addition, we applied the leave-one-out technique to assess the importance of the attributes used in graph matching, highlighting the significance of each attribute and enhancing the interpretability of the model.

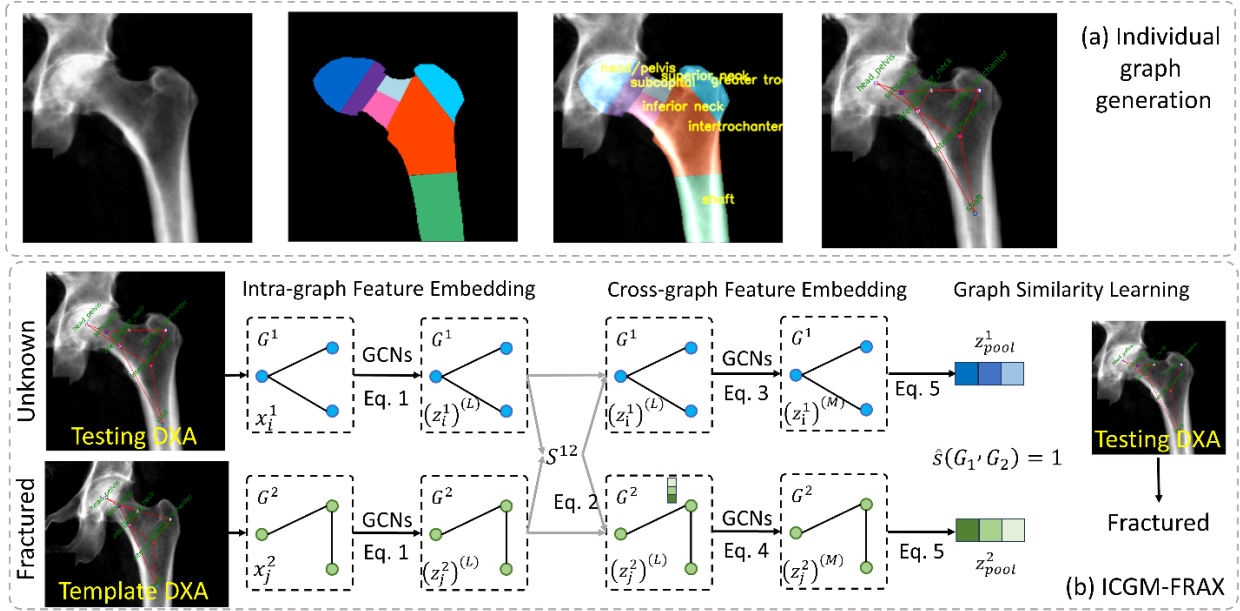

**Figure 1**. The workflow of the proposed ICGM-FRAX for hip fracture risk assessment: (a) The process of semantic graph generation using DXA. Starting with the raw DXA image, we label each RoI of the left femur. Based on the centroid of each RoI, we build a graph to connect the RoIs. (b) The process of hip fracture risk assessment by comparing one DXA scan from a subject with unknown fracture status and one DXA scan from a subject with a hip fracture. Using the procedure in (a), individual semantic graphs are created. By applying the proposed ICGM-FRAX, which includes an intra-graph feature embedding module, cross-graph feature embedding module, and graph similarity measurement module, the similarity between these two graphs is derived. If the two graphs are identical, the degree of similarity is defined as 1, and this indicates that the DXA scan with unknown fracture status has the same likelihood of hip fracture as the template DXA. In the ICGM-FRAX framework, majority voting determines the final prediction by comparing a test DXA with multiple template DXAs, assigning the most frequent classification (fractured or non-fractured) as the final fracture risk assessment result. Our code is publicly available at https://github.com/MIILab-MTU/ICGM_FRAX.

The major contribution of this paper is shown below:

1) We propose a novel method that defines the DXA-derived semantic image as a graph and applies a graph matching algorithm to predict hip fractures by comparing the similarity between graphs derived from different subjects.

2) The proposed ICGM-FRAX method demonstrates not only high performance but also high interpretability.

3) The proposed method can also be applied to other diagnostic tasks using medical images where the connectivity and relationships between semantic RoIs are crucial.

## 2. Related Work

### 2.1 Hip Fracture Risk Assessment using DXA

DXA has long been a standard imaging technique for measuring BMD, a key indicator of osteoporosis and fracture risk [6]. However, its ability to predict hip fractures specifically, while valuable, is limited when used in isolation. Recent advances have incorporated machine learning models and imaging techniques to improve the predictive power of DXA scans.

Several studies have demonstrated the potential of combining DXA images with advanced machine learning techniques for improved hip fracture risk assessment. One study developed a deep learning-based approach using DXA images of fallers and non-fallers, showing that integrating deep neural networks with both DXA images and clinical data resulted in an area under the receiver operating characteristic curve (AUROC) of 74.3% for fracture risk classification [7]. This approach highlights the importance of leveraging deep learning for feature extraction from DXA images, which could enhance the accuracy of hip fracture risk assessment, particularly in subjects with a history of falls.

In addition to machine learning models, novel imaging-based approaches have been explored to enhance DXA's predictive capability. One such method involves the use of two-dimensional finite element (FE) models derived from DXA scans. These models simulate the mechanical behavior of the femur during a sideways fall and calculate fracture risk based on principal strain criteria. A study comparing DXA-based 2D FE models with 3D models validated their effectiveness in predicting hip fracture risk, showing strong correlations (Spearman's $\rho = 0.66$, p < 0.001) between the 2D models and the 3D models, as well as with T-scores (Spearman's $\rho = -0.69$, p < 0.001) [8]. The integration of DXA scans with FE modeling presents a tool for personalized fracture risk assessment, offering more granular insights into bone quality and fracture susceptibility.

Additionally, advancements in automated tools have been developed to predict BMD and fracture risk using plain radiographs, which are often more accessible than DXA scans. One such tool demonstrated high accuracy in identifying osteoporosis and fracture risk from radiographs, achieving precision-recall area under the curve (PR AUC) scores of 0.89 for hip osteoporosis and 0.96 for high hip fracture risk [9]. This automated approach provides a potential alternative or supplement to DXA scans, especially in settings where access to DXA may be limited. In our previous study, a staged approach using uncertainty quantification determined when DXA features were needed for better prediction [10]. The staged model achieved the best performance with an AUC of 0.95 and suggested that 54.49% of patients did not need DXA scanning. This approach balanced accuracy and specificity, especially when DXA data were unavailable.

While DXA remains a critical tool in fracture risk assessment, combining it with advanced machine learning models, finite element analysis, and automated tools significantly enhances its predictive accuracy. These innovations improve clinical decision-making by providing more accurate and personalized hip fracture risk assessments, ultimately leading to better patient outcomes and more efficient healthcare delivery. Furthermore, it offers a promising new approach to enhance the accuracy of hip fracture risk assessment by capturing the complex spatial relationships between regions of interest in DXA images, which traditional methods may overlook.

**2.2 Graph Matching for Graph Similarity Learning**

Graph similarity learning, a critical task in various domains, has seen significant advancements with the integration of deep learning techniques. Traditional methods often relied on predefined similarity measures, which could be limited in capturing complex graph structures [11]. Recent approaches leverage deep learning to automatically learn similarity metrics, enhancing performance in tasks like classification, clustering, and retrieval.

One prominent category involves Graph Neural Networks (GNNs), which are adept at capturing local and global graph structures [12]. For instance, Siamese GNNs have been employed to learn embeddings for graph pairs, enabling effective similarity assessments [13]. Additionally, graph matching networks utilize attention mechanisms to align nodes between graphs, facilitating precise similarity computations [14]. Another approach is the development of deep graph kernels [12]. These methods learn kernel functions that measure graph similarity by mapping graphs into a feature space where similarities correspond to distances between embeddings. This technique combines the expressiveness of kernel methods with the representation learning capabilities of deep networks.

While existing deep learning-based graph similarity learning methods, such as Siamese GNNs and graph matching networks, have demonstrated effectiveness in various domains, they often face challenges in handling structured medical data, particularly in fracture risk assessment. The proposed ICGM-FRAX model introduces a novel approach tailored for hip fracture risk assessment on DXA scans. This model's novelty lies in its domain-specific adaptation of graph matching techniques, improving interpretability and accuracy in medical applications while outperforming traditional deep learning-based methods.

## 3. Methodology

This study introduces an image-level classification approach for assessing hip fracture risk using DXA images. The proposed ICGM-FRAX framework transforms hip fracture risk assessment into a problem of evaluating the similarity between a test DXA image and multiple template DXA images. The overall workflow is depicted in Figure 1, with further details provided in the following sections.

### 3.1 Modeling DXA in Graph

We annotate the contours of left femur using DXA images by LabelMe [15]. Then, we manually draw the RoIs of the femur, including femur head, subcapital, inferior neck, superior neck, intertrochanteric, greater trochanter and femur shaft if presented. Example of annotation is shown in Figure 1 (a). In detail, the first step in the process involves loading the necessary data, including the binary mask and JSON annotation file generated by LabelMe. The binary mask ensures that only relevant anatomical structures are considered, while the JSON file contains the polygon annotations that define the RoIs within the DXA image. Each RoI corresponds to a specific anatomical region, such as the femoral head or lesser trochanter.

**Feature extraction**: Once the RoIs are identified, their centroids are computed using image moments. A mask is generated for each RoI based on the polygon annotations, and the centroid coordinates $(x, y)$ are extracted, playing a crucial role in defining the spatial structure of the graph representation. Following centroid extraction, radiomics features [16,17] are computed from the cropped RoI image and its corresponding mask. These features capture texture, intensity, and shape-based characteristics of the bone regions, providing valuable insights for fracture risk assessment. Additionally, clinical features associated with the patient are integrated into the feature vector. If the cropped image is unavailable, a placeholder vector is assigned to maintain consistency across all RoIs. The extracted features and centroid information are then structured into a dictionary, which serves as input for graph construction.

**Graph construction**: Once the RoIs have been processed, the DXA image is converted into a graph representation, where each RoI serves as a *node*. The spatial relationships between these RoIs determine the graph's connectivity, indicating the anatomical structure is accurately represented. In this study, k-nearest neighbors (kNN) method is used to establish connectivity between nodes. Each RoI is linked to its $k$ closest neighbors based on the Euclidean distance between their centroids. kNN derived graph captures the local relationships between different anatomical regions. By adjusting the $k$ value, the level of connectivity can be controlled, balancing between sparse and overly dense graph structures.

Furthermore, to ensure all nodes are part of a single connected component, an iterative approach is applied. If any nodes remain disconnected, the number of nearest neighbors, i.e. $k$ or the distance threshold is adjusted until all nodes are connected. This step is essential to prevent isolated RoIs from being excluded from the analysis, ensuring that the entire anatomical structure contributes to hip fracture risk assessment.

At the end of this process, the DXA image is successfully transformed into a graph representation. In this graph, nodes correspond to RoIs, enriched with radiomics and clinical features, while edges define their spatial relationships based on the selected connectivity method. Formally, the derived graph is defined as $G = (V, E)$, where $V = \{V_1, V_2, \cdots V_n\}$ and $E = \{E_1, E_2, \cdots, E_{n_e}\}$, indicting the node set and edge set, respectively. $n$ represents the number of nodes, as well as the number of RoIs in the DXA image, and $n_e$ is the number of edges in the DXA derived graph.

### 3.2. Iterative Cross Graph Matching for Hip Fracture Risk Assessment using DXA

DXA derived structured graph serves as input for graph matching, where patient graphs are compared against known templates. By analyzing graph similarities, hip fracture risk can be effectively assessed, providing a novel and interpretable approach for hip fracture risk assessment. Formally, given two DXA-derived graphs, $G^1 = (V^1, E^1)$ and $G^2 = (V^2, E^2)$, the proposed **I**terative **C**ross **G**raph **M**atching for Hip Fracture **R**isk **A**ssessment (**ICGM-FRAX**) aims to calculate the similarity between $G^1$ and $G^2$. If the DXA corresponding to $G^1$ comes from a normal patient, and $G^2$ comes from a hip-fractured patient, then the similarity between $G^1$ and $G^2$ is zero; otherwise, if both $G^1$ and $G^2$ are derived from DXA images of subjects within the same group, the similarity between $G^1$ and $G^2$ is 1. The overall graph matching network contains the following five modules, as shown in Figure 1.

1) *Extracting features for nodes in individual graphs*. To construct individual graphs, a processing algorithm is applied to generate a graph representation for each DXA image. In this representation, each node corresponds to a RoI within the femur. For each node, we extract pixel-based radiomics features while clinical attributes are assigned based on patient information. These combined features are represented as $x_i^g \in \mathbb{R}^d$, where $i \in \{1, 2, \cdots, n\}$ indicates the node index in the DXA-derived graph, $d$ denotes the dimensionality of the extracted feature set, and $g \in \{1, 2\}$ identifies the graph being compared. Specifically, we extract $d = 130$ features, as shown in Table 1.

**Table 1.** Feature for each node (RoI) in the DXA-derived graph

| Index | Feature explanation |
|---|---|
| 1-5 | These radiomics features include the mean, minimum, and maximum intensity values of the RoI, along with the voxel count and volume, which provide insights into the region's brightness, size, and spatial extent. |
| 6-16 | Shape-based radiomics features describe the geometric properties of the RoI, including elongation, axis lengths, maximum 2D diameters, mesh area, minor axis length, sphericity, surface area, surface-to-area ratio, and pixel-based area, providing insights into the region's size, shape, and structural complexity. |
| 17-34 | First-order radiomics features, which capture the intensity distribution within the RoI, including percentile values, energy, entropy, statistical dispersion (interquartile range, variance, skewness, kurtosis), and measures of central tendency (mean, median), providing quantitative insights into image texture and brightness variations. |
| 35-58 | Gray Level Co-occurrence Matrix (GLCM) features, which quantify textural patterns within the RoI by measuring spatial relationships between pixel intensities, capturing properties such as contrast, correlation, entropy, homogeneity, and cluster tendencies to describe image texture and structural complexity. |

| 59-73 | Gray Level Dependence Matrix (GLDM) features, which characterize texture by measuring the distribution and variance of dependent gray levels, capturing patterns related to dependence size, gray-level uniformity, and emphasis on high or low gray-level intensities within the RoI. |
|---|---|
| 74-89 | Gray Level Run Length Matrix (GLRLM) features quantify texture patterns by analyzing the distribution of consecutive pixels with the same intensity, capturing variations in run length, gray-level uniformity, and emphasis on short or long runs within the RoI. |
| 90-105 | Gray Level Size Zone Matrix (GLSZM) features, which evaluate textural patterns based on the size and distribution of homogeneous zones with the same gray level, capturing variations in zone size, gray-level uniformity, and emphasis on large or small areas within the RoI. |
| 106-110 | Neighborhood Gray Tone Difference Matrix (NGTDM) features, which assess texture by analyzing the variation in gray-tone differences between neighboring pixels, capturing properties such as busyness, coarseness, complexity, contrast, and strength within the RoI. |
| 111-121 | Clinical features, including the demographic, lifestyle, and health-related attributes, such as age, sex, height, weight, household size, smoking and alcohol habits, diet, dietary changes, falls in the last year, and fractures or broken bones in the past five years. |
| 122-130 | Features which are related to bone mineral density (BMD) and bone mineral content (BMC) measurements of the femur and pelvis, with specific values for the left femur (denoted by "L") and T-scores, reflecting the density of the femoral neck, total femur, trochanter, and wards area, as well as the pelvis BMC. |

2) *Intra-graph Feature Embedding*. The feature embedding module utilizes graph convolutional networks (GCNs) and multi-layer perceptron (MLP) to capture a comprehensive representation of each arterial segment. In the context of a DXA derived graph $G$, the intra-graph feature embedding for node $V_i$ is represented as in Eq. 1.

$$z_i^{(l+1)} = f_{intra\_emb}^{(l)}\left(\left[\sum_{j \in E_i} z_j^{(l)}, z_i^{(l)}\right]\right) \quad (1)$$

where $l \in \{1, \cdots, L\}$ denotes the GCN layer index; $f_{intra\_emb}^{(l)}$ is the MLP for the $l$-th layer; $E_i$ contains the edges connected with node $V_i$; the summation of $\sum_{j \in E_i} \cdot$ represents the element-wise aggregation of features from adjacent nodes, where the adjacency is determined by the adjacency matrix of $G$. If $l = 1$, then the node feature $z_i^1 = x_i \in \mathbb{R}^d$ for node $V_i$ in graph $G$. The square brackets $[\cdot]$ signify feature concatenation. The equation models the message passing and aggregation process within the GCN for feature embedding. Then, Eq. 1 represents the message passing and aggregation for feature embedding using GCN in graph $G$. As a result, the final intra-graph node embedding is denoted as $z_i^{(L)}$ for node $V_i$ in graph $G$.

The described embedding approach encodes the structural similarities between two graphs by computing node-to-node affinities in the embedding space. For two graphs $G^1$ and $G^2$, the intra-graph node similarity is determined using the weighted dot product of the updated feature embeddings for each pair of nodes. A Sinkhorn operator is then used to convert the resulting node similarity matrix into a doubly-stochastic matrix [18], as shown in Eq. 2.

$$S_{ij} = Sinkhorn\left(\exp\left(\frac{(z_i^1)^{(L)} \cdot A_{intra} \cdot (z_j^2)^{(L)}}{\sqrt{d_{intra}}}\right)\right) \quad (2)$$

where $i$ and $j$ are the node indices in $G^1$ and $G^2$; $d_{intra}$ is the feature dimension, and $A_{intra} \in \mathbb{R}^{d_{intra} \times d_{intra}}$ consists of learnable parameters for node affinity. The exponential function ensures the non-negativity of the node-to-node affinity, which is necessary for the Sinkhorn algorithm [18]. Eq. 2 integrates GCN-based feature embedding with Sinkhorn's algorithm to enable graph matching and affinity assignment between nodes across two graphs [19].

3) *Cross-graph Feature Embedding in Graph Pairs*. Cross-graph feature embedding is a critical process for improving the reliability of node correspondences between graphs. Without interactive aggregation across graphs, direct node-to-node matching often lacks robustness [20]. This method involves encoding nodes and edges from two distinct graphs, aiding in the enhancement of graph matching performance [21]. For two graphs $G^1$ and $G^2$, the cross-graph feature embedding for node $V_i$ in $G^1$, considering all nodes in $G^1$ and $G^2$, is defined as shown in Eq. 3.

$$\left(z_i^1\right)^{(m+1)} = f_{cross\_emb}^{(m)}\left(\left[\sum_{j=1}^{n_2} S_{ij} \cdot \left(z_j^2\right)^{(m)}, \left(z_i^1\right)^{(m)}\right]\right) \tag{3}$$

In the cross-graph embedding process, $f_{cross\_emb}^{(m)}$ represents the MLP used for the $m$-th layer of the cross-graph embedding module, where $m \in \{1, \cdots, M\}$. When $m = 1$, $\left(z_i^1\right)^{(1)}$ corresponds to $\left(z_i^1\right)^{(L)}$, the feature embedding from the final layer of the intra-graph feature embedding. The cross-graph feature embedding is applied for $M$ times to extract hierarchical features across graphs and perform non-linear transformations.

Symmetrically, for node $V_j$ in $G^2$, the updated cross-graph feature embedding is calculated as shown in Eq. 4.

$$\left(z_j^2\right)^{(m+1)} = f_{cross\_emb}^{(m)}\left(\left[\sum_{i=1}^{n_1} S_{ij} \cdot \left(z_i^1\right)^{(m)}, \left(z_j^2\right)^{(m)}\right]\right) \tag{4}$$

As a result, after applying the intra-graph and cross-graph feature embedding, the features for node $V_i$ are denoted as $\left(z_i^g\right)^{(M)} \in \mathbb{R}^{d_{cross}}$, s.t. $g \in \{1,2\}$ and $d_{cross}$ indicates the dimension of the feature representation.

4) *Global feature extraction*. After the feature embedding process, graph pooling is applied to generate the average representation for each graph. Formally, the features for graph $G^g$ after applying the graph average pooling operation are calculated in Eq. 5.

$$z_{pool}^g = \frac{1}{n}\sum_{i=1}^{n}\left(z_i^g\right)^{(M)} \tag{5}$$

where $z_{pool}^g \in \mathbb{R}^{d_{cross}}$ is the average feature vector representing the entire graph $G^g$. In addition, the pooling operation aggregates node features by computing the mean across all nodes in the graph, effectively creating a fixed-size graph representation from varying node features.

5) *Graph similarity measurement*. Specifically, the pooled feature embeddings for graph $G^1$ and $G^2$ are used to compute the similarity between the two graphs. In particular, the cosine similarity is calculated between the averaged embeddings of both graphs, serving as a measure of their similarity, which is denoted in Eq. 6. This approach is both efficient and effective when the node features carry significant information.

$$\hat{s}(G_1, G_2) = \frac{z_{pool}^1 \cdot z_{pool}^2}{\|z_{pool}^1\| \|z_{pool}^2\|} \tag{6}$$

where $z_{pool}^1$ and $z_{pool}^2$ are the averaged feature embeddings (after graph pooling) for the two graphs $G^1$ and $G^2$, respectively. $\hat{s} \in [0,1]$ indicates the predicted similarly between $G^1$ and $G^2$.

### 3.3. Model Training and Testing

The dataset is divided into three subsets: a training set, a testing set, and a template set, denoted as $D_{train}$, $D_{test}$ and $D_{temp}$, respectively.

**Model training:** During model training, two separate graphs with an equal number of RoIs in the femur were randomly chosen. At each training step, a batch of graph pairs was generated randomly to speed up the training process and reduce the risk of overfitting [22]. The ICGM-FRAX is trained using Mean Squared Error (MSE) as the loss function, as shown in Eq. 7.

$$L = \frac{1}{B} \sum_{b=1}^{B} \left( \hat{s}(G_1^b, G_2^b) - s(G_1^b, G_2^b) \right)^2 \tag{7}$$

where $B$ indicates the batch size and $b \in \{1,2,\cdots,B\}$ indicates the index of the training graph pair in the batch. The ground truth $s(G_1^b, G_2^b)$ is derived from the labels of the DXA images in $D_{train}$, where the similarity score is set to 1 if both graphs correspond to hip-fractured subjects or both graphs correspond to non-hip-fractured subjects, and 0 otherwise. The objective is to minimize the discrepancy between the predicted similarity score and the ground truth similarity score. The training algorithm is defined in Algorithm 1.

---

**Algorithm 1**. Training procedure of the proposed ICGM-FRAX

**Input**:

- $D_{train} = \{G_1^{train}, G_2^{train}, \cdots, G_{n_{tr}}^{train}\}$: $n_{tr}$ labeled DXA derived graphs for training.

**Output**:

- Trained ICGM

For $training\ step \in \{1, \cdots, N\}$ do:

   1. Randomly select two individual graphs $G_i^{train}$ and $G_j^{train}$;

   2. Extract features for each RoI in $G_i^{train}$ and $G_j^{train}$;

   3. Perform intra-graph feature embedding for $G_i^{train}$ and $G_j^{train}$ using Eqs. 1 and 2;

   4. Perform cross-graph feature embedding between $G_i^{train}$ and $G_j^{train}$ using Eqs. 3 and 4;

   5. Calculate the graph similarity between $G_i^{train}$ and $G_j^{train}$ using Eqs. 5 and 6;

   6. Optimize ICGM using the objective function defined in Eq. 7.

**Testing**: During testing, each graph from the testing set is compared against all graphs in the template set to compute the similarity score. If the similarity between a test graph and any template graph exceeds a predefined threshold, that match is retained. After iterating through all the graphs in the template set, we select graphs with a similarity greater than the threshold ($\theta$) and perform majority voting to generate the final prediction. If the majority of matching graphs from the template set belong to the hip-fracture group, the test graph is predicted as belonging to the hip-fracture group; otherwise, it is classified as a non-hip-fracture subject. The testing algorithm is illustrated in Algorithm 2.

---

**Algorithm 2**. Testing procedure of the proposed ICGM-FRAX

**Input**:

- $G^{test}$: unlabeled DXA derived graph for testing.
- $D_{temp} = \{G_1^{temp}, G_2^{temp}, \cdots, G_{n_{tp}}^{temp}\}$: $n_{tp}$ labeled DXA derived graphs as template.
- ICGM: trained ICGM-FRAX model
- $\theta$: threshold to accept the graph matching result

**Output**:

- hip fracture risk assessment result for $G^{test}$

  For each individual graph $G_j^{temp} \in D_{temp}$:

  1. Calculate the similarity between $G^{test}$ and $G_j^{temp}$ using trained ICGM, as $\hat{s}\left(G^{test}, G_j^{temp}\right)$

  2. If $\hat{s}\left(G_i^{test}, G_j^{temp}\right) > \theta$ then accept the graph matching results for majority voting

  3. Assign labels for $G^{test}$ according to major voting among $G_j^{temp}, j \in \{1, \cdots, n_{tp}\}$

---

### 3.4. Performance Evaluation

The performance of the model is evaluated using standard classification metrics, as the task of graph matching is framed as a binary classification problem where each graph is categorized as either "hip-fracture" or "non-hip-fracture." The key evaluation metrics used to assess the model's performance include accuracy (ACC), sensitivity (SN), specificity (SP), and F1 score.

In detail, Accuracy (ACC) measures the overall correctness of the model's predictions, calculated as the ratio of correctly classified instances to the total number of instances: $ACC = \frac{TP+TN}{TP+TN+FP+FN}$, where $TP$ is true positive, $TN$ is true negative, $FP$ is false positive, and $FN$ is false negative. Sensitivity quantifies the model's ability to correctly identify positive instances (hip-fracture cases). It is defined as the ratio of true positives to the total number of actual positives as $SN = \frac{TP}{TP+FN}$. Specificity measures the model's ability to correctly identify negative instances (non-hip-fracture cases). It is defined as the ratio of true negatives to the total number of actual negatives, as $SP = \frac{TN}{TN+FP}$. F1 Score provides a balance between precision and recall, especially useful in imbalanced datasets, defined as $F1 = 2 \times \frac{TP}{2 \times TP+FP+FN}$.

### 3.5. Interpretability of ICGM-FRAX

Given that the dataset is imbalanced, with a smaller proportion of hip-fracture subjects (positive class) compared to non-hip-fractured subjects (negative class), we emphasize the model's sensitivity during this analysis. Sensitivity, also known as recall, measures the ability of the model to correctly identify positive instances, i.e., hip-fracture subjects. In imbalanced datasets, where the number of negative samples significantly outweighs the positive ones, traditional performance metrics like ACC may not fully capture the model's effectiveness in detecting the minority class. A high accuracy might be achieved by the model simply predicting the majority class, but this would result in a poor ability to identify true positive cases. Therefore, by focusing on SN, we ensure that the model's performance is evaluated based on its capability to correctly identify hip-fracture subjects, which is critical for clinical applications where missing a hip-fracture diagnosis could have serious consequences.

To interpret the importance of individual features in the ICGM-FRAX model, we conduct a sensitivity analysis by systematically setting one feature to zero and feeding the modified features into the trained model. By observing the resulting drop in sensitivity, we can infer the significance of each feature. If the sensitivity decreases significantly after removing a particular feature, it suggests that the feature plays an important role in the model's decision-making process. This feature interpretation algorithm helps identify which features most strongly influence the model's ability to correctly classify hip-fracture cases.

### 4. Experimental Results and Analysis

### 4.1 Materials and Enrolled Subjects

The study cohort comprised 547 subjects, with 94 individuals who experienced hip fractures from UKBiobank dataset. Similarity to our previous publication [10], subjects who did not undergo DXA scanning were excluded based on the absence of significant risk factors, such as younger age, no history of prior fractures, lack of clinical risk factors for osteoporosis (e.g., smoking history, excessive alcohol consumption), and initial clinical assessments indicating a low fracture risk.

### 4.2 Implementation Details

We implemented our proposed ICGM-FRAX using PyTorch and NetworkX on a workstation with an NVIDIA RTX 4090 GPU. The threshold $\theta$ used in Algorithm 2 for model testing was empirically set to 0.5. For the 547 enrolled DXA images in the UK Biobank, we applied stratified sampling, with 10% of the DXA images selected as the template set ($D_{temp}$ in Algorithm 2). For the remaining 90% of the DXA images, we applied stratified sampling to generate the training set ($D_{train}$ in Algorithm 2) and testing set ($D_{test}$ in Algorithm 2), with 80% and 20% of the samples, respectively. We employed the Adam optimizer to fine-tune the model and set the batch size B in Eq. 7 to 64. Due to the limited availability of training samples, we repeated our experiments 10 times and reported the average performance.

### 4.3 Experimental Results of ICGM-FRAX

The best performance of the proposed ICGM-FRAX was achieved with the following settings: the dimensions of the intra-graph embedding, $d_{intra}$ and cross-graph feature embedding, $d_{cross}$, were set as 256. The number of GCN layers, i.e. $L$ in Eq. 1, was set as 5. And the number of cross-graph feature embedding, i.e. $M$ in Eq. 4, was set as 3. To compare performance, we set several baseline models and compared them with our previous staged model with ensemble [10]. For the baseline models, we employed transfer learning using ResNet18, ResNet50, and ResNet152 as the backbone to extract imaging features from the entire DXA images, with or without clinical features. The experimental results and comparisons are shown in Table 2.

**Table 2**. Performance comparison of the proposed ICGM-FRAX and baseline models. Mean performance ± standard deviation across 10 repeated experiments are reported.

| Model | ACC | F1 | SN | SP |
|---|---|---|---|---|
| ResNet18 with Clinical features | 0.7155± 0.0149 | 0.2496± 0.0198 | 0.2737± 0.0222 | 0.8077± 0.0174 |
| ResNet50 with Clinical features | 0.7782± 0.0107 | 0.2323± 0.0386 | 0.1947± 0.0355 | 0.9000± 0.0096 |
| ResNet152 with Clinical features | 0.7755± 0.0086 | 0.3777± 0.0229 | 0.3947± 0.0277 | 0.8549± 0.0087 |
| ResNet18 without Clinical features | 0.7518± 0.0136 | 0.2796± 0.0193 | 0.2789± 0.0254 | 0.8505± 0.0181 |
| ResNet50 without Clinical features | 0.7600± 0.0077 | 0.1577± 0.0401 | 0.1316± 0.0372 | 0.8912± 0.0121 |
| ResNet152 without Clinical features | 0.0771± 0.0058 | 0.3572± 0.0055 | 0.3684± 0.0000 | 0.8549± 0.0070 |
| ICGM-FRAX | 0.9970± 0.0068 | 0.9932± 0.0162 | 0.9869± 0.0310 | 1.0000± 0.0000 |

The experimental results demonstrate the superiority of the proposed ICGM-FRAX model over traditional deep learning approaches for hip fracture risk assessment. Compared to baseline models, which employ ResNet architectures with and without clinical features, ICGM-FRAX achieved significantly higher accuracy, F1-score, sensitivity, and specificity.

One of the key advantages of ICGM-FRAX is its ability to integrate both imaging and clinical data within a graph-based framework. The baseline models, even when incorporating clinical features, struggled to achieve high sensitivity, with the best-performing ResNet model (ResNet152 with clinical features) achieving a sensitivity of only 0.3947. In contrast, ICGM-FRAX exhibited near-perfect sensitivity (0.9869), ensuring that hip fracture cases were accurately identified. This improvement is crucial for clinical applications where false negatives could have severe consequences for patient outcomes.

### 4.4 Hyperparameter Setting and Ablation Study

We further fine-tuned the proposed ICGM-FRAX by experimenting with different values for key hyperparameters. Specifically, we set the dimensions of the intra-graph embedding ($d_{intra}$) and cross-graph feature embedding ($d_{cross}$) to $\{32, 64, 128, 256\}$. The number of GCN layers ($L$, as defined in Eq. 1) was set to $\{2,3,4,5\}$, while the number of cross-graph feature embedding layers (M, as defined in Eq. 4) was set to $\{1,2,3,4\}$. We reported the sensitivity among different settings, as shown in Figure 2.

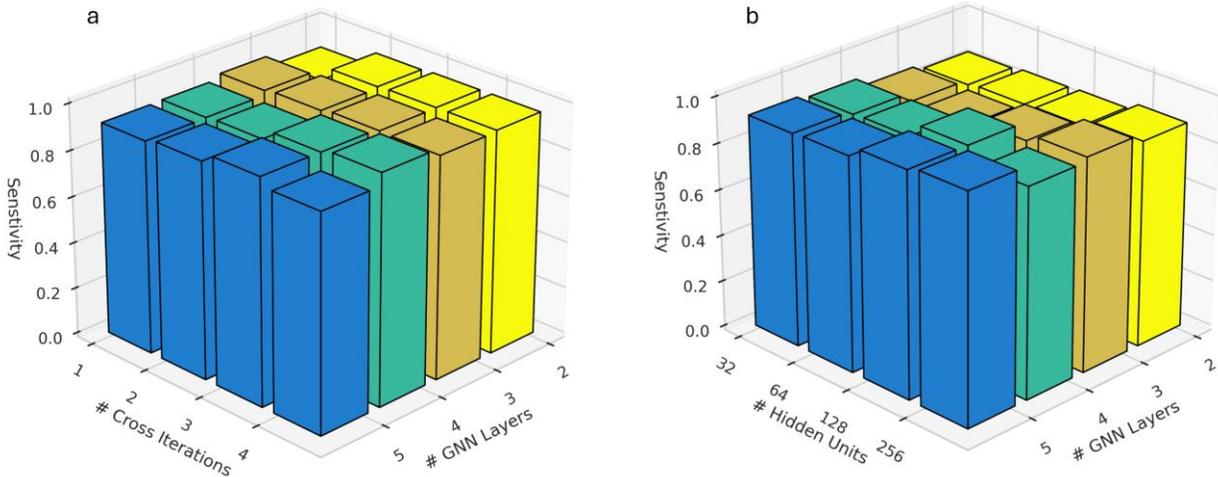

**Figure 2**. Hyperparameter analysis of ICGM-FRAX on UKBiobank dataset under different settings. (a) The number of cross iterations indicate $M$ in Eq. 4 and the number of GNN layers indicates $L$ in Eq. 1; (b) The number of hidden unites indicate $d_{intra}$ and $d_{cross}$. We set the $d_{intra} = d_{cross}$ in this experiment and applied the cross-graph feature embedding to each experiment.

The hyperparameter tuning experiments (Figure 2) provide valuable insights into the impact of different settings on the sensitivity of ICGM-FRAX. In Figure 2 (a), we observe that increasing the number of cross iterations ($M$) improves sensitivity, especially when combined with a higher number of GCN layers ($L$). The performance stabilizes when $M \geq 3$ and $L \geq 3$, suggesting that deeper graph neural networks and additional cross-graph embedding iterations contribute to better feature extraction and classification. However, beyond a certain threshold ($M = 4, L = 5$), the improvements plateau, indicating that further increasing these hyperparameters does not significantly enhance performance.

Figure 2 (b) illustrates the effect of varying the hidden unit dimensions ($d_{intra}$ and $d_{cross}$). We observe that larger hidden unit dimensions (e.g., 128 and 256) lead to a noticeable performance gain compared to smaller dimensions (32 and 64), likely due to the increased capacity to capture complex structural patterns. However, the performance gain between 128 and 256 is marginal, indicating that a dimension of 128 might provide a good balance between performance and computational cost.

These findings highlight the importance of selecting appropriate hyperparameters to balance model complexity and performance. In particular, choosing $M = 3\ or\ 4$, $L = 3\ or\ 4$, and $d_{intra} = d_{cross} = 128$ appears to offer an optimal trade-off between sensitivity and computational efficiency.

For the ablation study, we removed the cross-graph feature embedding and directly employed the features after applying the intra-graph embedding for graph similarity learning. The comparison is shown in Table 3.

**Table 3**. Ablation study on cross-graph feature embedding. We tested model performance with different numbers of GNN layers ($L$) and varying numbers of hidden units in GNN ($D_{intra}$). The average performance and standard deviation are reported.

| Model | ACC | F1 | SN | SP | $L$ | $D_{intra}$ |
|---|---|---|---|---|---|---|
| ICGM w/o cross graph embedding | 0.9686±0.0074 | 0.9122±0.0316 | 0.8400±0.0536 | 1.0000±0.0000 | 2 | 32 |
| | 0.9696±0.0106 | 0.9194±0.0276 | 0.8520±0.0474 | 1.0000±0.0000 | 2 | 64 |
| | 0.9747±0.0136 | 0.9316±0.0373 | 0.8740±0.0667 | 1.0000±0.0000 | 2 | 128 |
| | 0.9818±0.0149 | 0.9487±0.0426 | 0.9052±0.0767 | 1.0000±0.0000 | 2 | 256 |
| | 0.9666±0.0095 | 0.9051±0.0395 | 0.8289±0.0661 | 1.0000±0.0000 | 3 | 32 |
| | 0.9757±0.0136 | 0.9366±0.0359 | 0.8827±0.0630 | 1.0000±0.0000 | 3 | 64 |
| | 0.9858±0.0097 | 0.9626±0.0306 | 0.9294±0.0554 | 1.0000±0.0000 | 3 | 128 |
| | 0.9797±0.0157 | 0.9391±0.0585 | 0.8903±0.1010 | 1.0000±0.0000 | 3 | 256 |
| | 0.9696±0.0095 | 0.9148±0.0333 | 0.8445±0.0574 | 1.0000±0.0000 | 4 | 32 |
| | 0.9777±0.0149 | 0.9334±0.0553 | 0.8796±0.0948 | 1.0000±0.0000 | 4 | 64 |
| | 0.9898±0.0134 | 0.9743±0.0335 | 0.9518±0.0619 | 1.0000±0.0000 | 4 | 128 |
| | 0.9838±0.0127 | 0.9523±0.0464 | 0.9123±0.0818 | 1.0000±0.0000 | 4 | 256 |
| | 0.9696±0.0116 | 0.9158±0.0375 | 0.8468±0.0644 | 1.0000±0.0000 | 5 | 32 |
| | 0.9676±0.0104 | 0.9113±0.0305 | 0.8384±0.0529 | 1.0000±0.0000 | 5 | 64 |
| | 0.9868±0.0107 | 0.9682±0.0238 | 0.9394±0.0452 | 1.0000±0.0000 | 5 | 128 |
| | 0.9898±0.0082 | 0.9743±0.0206 | 0.9507±0.0393 | 1.0000±0.0000 | 5 | 256 |
| ICGM | **0.9970± 0.0068** | **0.9932± 0.0162** | 0.9869± 0.0310 | 1.0000± 0.0000 | 5 | 256 |

The ablation study results shown in Table 3 demonstrate the significance of the cross-graph feature embedding in enhancing the predictive performance of ICGM-FRAX. When the cross-graph feature embedding is removed, we observe a noticeable decline in all evaluation metrics, particularly in SN and F1.

Without cross-graph embedding, the model achieves a maximum ACC of 0.9898 and F1-score of 0.9743, compared to 0.9970 ACC and 0.9932 F1-score when the embedding is included. More importantly, the SN, which measures the model's ability to correctly identify hip fracture cases, is significantly lower in the ablation models. The highest SN without cross-graph embedding is 0.9518, whereas the full model achieves 0.9869, suggesting that cross-graph embedding plays a crucial role in capturing discriminative features for hip fracture risk assessment.

The effect of increasing the number of GCN layers ($L$) and hidden unit dimensions ($d_{intra}$) also becomes evident. Across different configurations, higher values of $L$ and $d_{intra}$ tend to improve performance, though the improvements plateau beyond $L = 4$ and $d_{intra} = 128$. This indicates that deeper networks and richer feature embeddings enhance model performance but only up to an optimal point, after which the gains diminish.

The complete ICGM-FRAX model, which incorporates cross-graph feature embedding, five GCN layers ($L = 5$), and 256 hidden units ($d_{intra} = 256$), achieves the best overall performance. This confirms that leveraging cross-graph embeddings significantly enhances feature learning and classification accuracy, particularly in detecting hip fracture cases.

Overall, the study of ablation highlights that removing cross-graph embedding leads to suboptimal performance, particularly in sensitivity. Thus, it is a crucial component for achieving high predictive accuracy in ICGM-FRAX.

**4.5 Feature Interpretation**

In this study, we designed 130 hand-craft features, as shown in Table 1. We also set the graph matching result acceptance threshold $\theta$ as 0.8 to interpret the feature importance using the algorithm defined in section 3.5. In our experiment, 98 cases were split into the testing set as $D_{test}$ and 55 cases were split into the template set as $D_{temp}$. During the feature importance explanation, we generate 3036 graph matching pairs when the testing graph and template graph contain the same number of RoIs. According to the SN drop, we further visualize the top 30 important features, as shown in Figure 3.

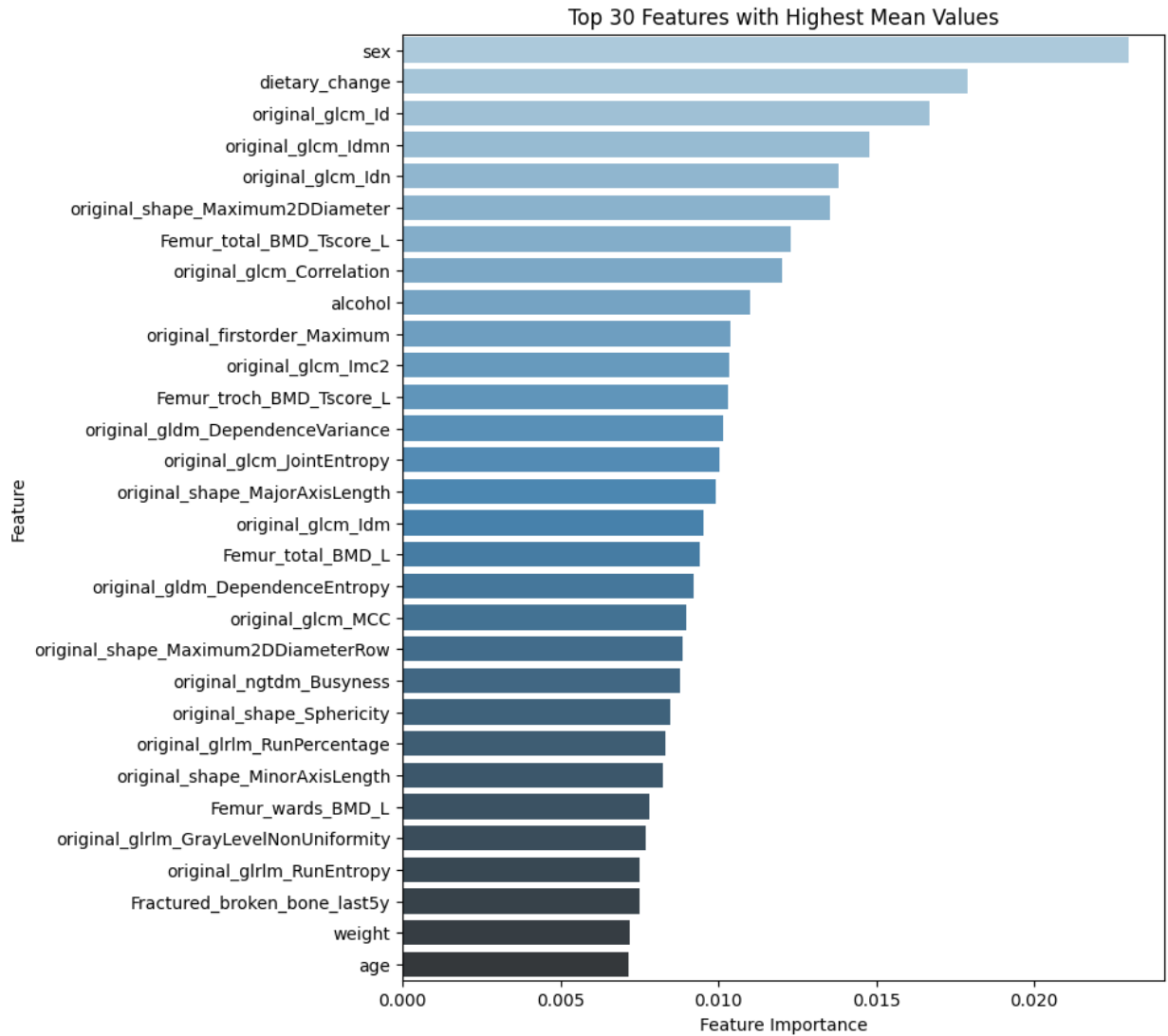

**Figure 3**. Feature importance ranking for classifying the DXA-derived graph for hip fracture risk assessment using the proposed ICGM-FRAX. The importance of the features was surrogated by the sensitivity drop after setting the corresponding features to zero during the graph matching. The horizontal axis represents the sensitivity drops, while the vertical axis represents the feature names.

The feature importance analysis provides valuable insights into the key predictors of hip fracture risk based on DXA-derived graph representations. As illustrated in Figure 3, sex emerged as the most influential feature, followed by dietary change and multiple radiomic and clinical variables.

Among the radiomic features, gray-level co-occurrence matrix (GLCM) and gray-level dependence matrix (GLDM) exhibited strong predictive value, particularly Id, Idmn, and Joint Entropy. These features quantify textural properties of bone microarchitecture, suggesting that subtle variations in bone texture contribute significantly to fracture risk. Additionally, Maximum 2D Diameter and Major/Minor Axis Length indicate the relevance of structural bone morphology in assessing fracture susceptibility.

Clinical variables such as total femoral BMD T-score and alcohol consumption also ranked among the top predictors. A lower BMD T-score is a well-established risk factor for fractures, supporting the reliability of our model in identifying clinically meaningful features. The influence of alcohol consumption may be linked to its impact on bone metabolism and fall risk, reinforcing its role as a modifiable risk factor.

Interestingly, traditional risk indicators like age and weight showed lower feature importance compared to radiomic descriptors. This finding suggests that radiomic and graph-based features may offer additional discriminatory power beyond conventional demographic and clinical parameters. Furthermore, the presence of fractured broken bone during the last 5 years (Fractured_broken_bone_last5y) among the top 30 features highlights the importance of prior fracture history in risk stratification.

Hence, the feature ranking underscores the advantage of integrating radiomic, clinical, and lifestyle factors to enhance hip fracture risk assessment. The proposed ICGM-FRAX framework effectively leverages these diverse data sources to improve fracture risk assessment, offering a more nuanced and personalized approach to patient evaluation.

We also reported the performance of ICGM-FRAX using top $K \in \{5, 10, 15, \cdots, 130\}$ most important features to perform hip fracture risk assessment. The performance is shown in Figure 4.

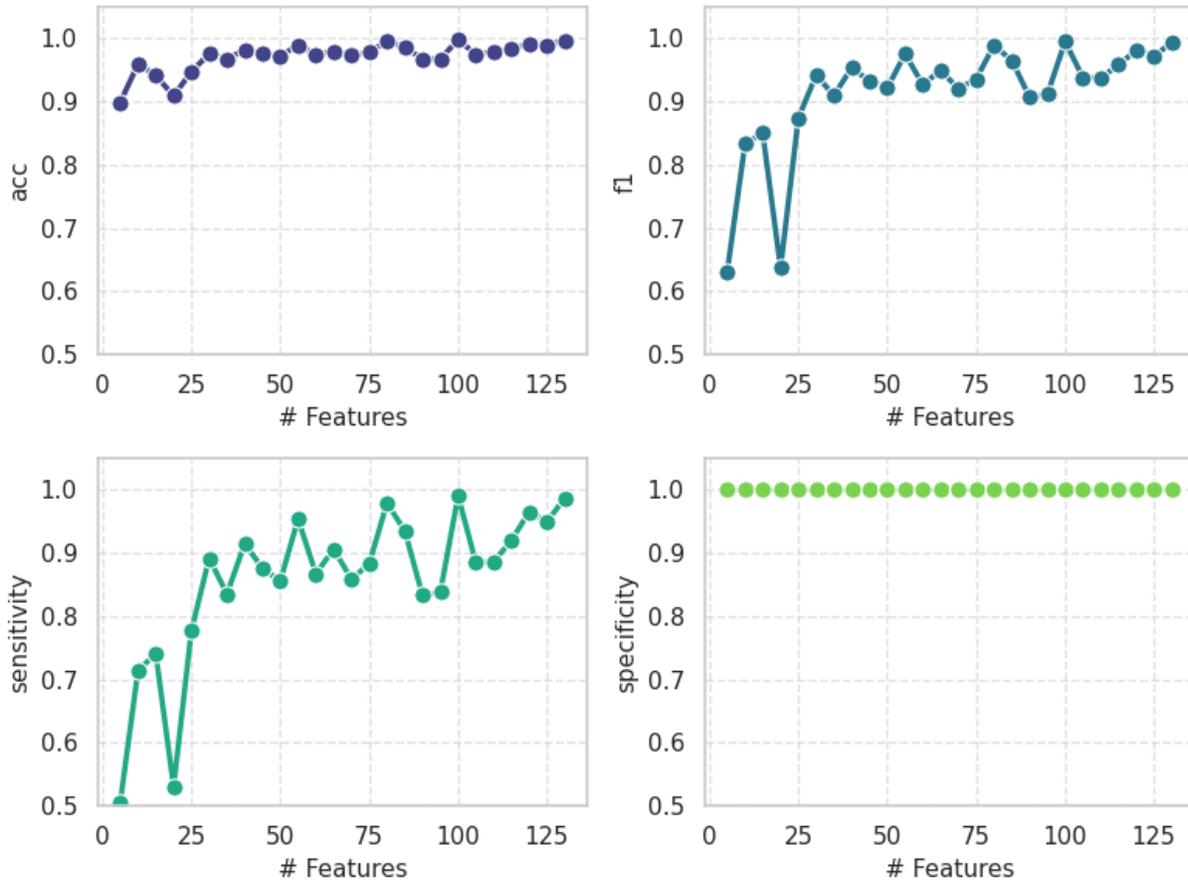

**Figure 4**. Plots showing ACC, F1, SN, and SP of our proposed ICGM-FRAX using different numbers of the top-*K* features. We reported the average performance based on the average performance among 10 repeated experiments.

The performance evaluation of ICGM-FRAX using different subsets of top-*K* features (Figure 4) provides insights into the trade-off between model complexity and predictive accuracy. The ACC and F1 plots indicate that performance stabilizes after incorporating approximately 20-30 features, suggesting that a relatively small subset of the most important features is sufficient to achieve high predictive performance. This is an encouraging finding, as it implies that feature selection can reduce computational complexity without compromising accuracy.

The SN plot shows an increasing trend as more features are incorporated, with notable fluctuations when fewer than 30 features are used. This indicates that certain key features are essential for improving the model's ability to correctly identify positive cases (i.e., fracture-prone individuals). However, beyond 75-100 features, SN stabilizes, reinforcing the idea that feature redundancy might not significantly contribute to further improvements.

In contrast, the SP remains consistently high across all feature subsets, hovering close to 1.0. This suggests that the model effectively maintains a low false-positive rate regardless of the number of selected features. The stability of specificity highlights the robustness of the model in correctly identifying non-fracture cases, which is crucial for reducing unnecessary clinical interventions.

These results indicate that while increasing the number of features can enhance SN, the marginal gains become minimal after a certain threshold. A carefully selected subset of approximately 30-50 features appears to provide an optimal balance between ACC, SN, and computational efficiency.

### 4.6 Graph Connectivity and Graph Generation

In our implementation, we adopt the kNN method to connect different RoIs in the femur image. The graph is constructed by ensuring each node is connected to its k nearest neighbors. This method leverages the power of proximity-based relationships by connecting a point to its closest $k$ neighbors in terms of Euclidean distance. By considering only a fixed number of neighbors, the kNN method controls the complexity of the graph, ensuring that each point is only influenced by its immediate adjacent RoIs. This method is particularly useful in identifying local clusters and capturing intricate neighborhood structures within the DXA image data, thus aiding in the identification of potential fracture risks based on spatial correlations between regions of interest.

Besides kNN, two other methods are still available.

1) *Delaunay triangulation method* creates the structured graph by connecting points to form triangles, ensuring that no point lies inside the circumcircle of any triangle in the mesh. This method guarantees that the resulting graph captures the spatial distribution of points in a way that maximizes the minimum angle of all the triangles, preventing thin, elongated triangles. Delaunay triangulation is particularly advantageous for generating a well-connected graph structure, where each point is linked to others in a manner that reflects the overall geometric arrangement. In the context of hip fracture risk assessment, Delaunay triangulation allows for capturing more complex, global relationships between RoIs, thereby improving the ability to model structural interactions in the DXA images.

2) *Distance-based threshold method* involves generating a graph by connecting nodes based on the centroids of each RoIs. The Euclidean distance is calculated, and an edge is formed between the points if their distance is below a predefined threshold. This approach is effective in capturing local relationships between points that are spatially close to each other. Distance threshold method highlights the most relevant spatial relationships in the image data. This method's simplicity and efficiency make it suitable for capturing neighborhood-based patterns without requiring complex machine learning models.

However, due to the simplicity of the femur RoIs, using the kNN method has already achieved high performance. Replacing the graph generation algorithm did not improve performance. We provide performance differences using different edge generation methods in the supplementary results, as shown in Table S1.

### 5. Conclusion

In this study, we propose a novel method, ICGM-FRAX, which defines the DXA-derived semantic image as a graph and applies a graph matching algorithm to predict hip fractures. This approach enables the model to compare the similarity between graphs derived from different subjects, effectively capturing the connectivity and relationships between RoIs of the femur. The ICGM-FRAX method demonstrates not only high performance but also exceptional interpretability. It achieves remarkable ACC of 99.7% and SN of 98.7% with minimal false positives, outperforming several baseline models that employed traditional transfer learning with ResNet backbones. The feature importance analysis further supports this by revealing that key features, such as sex, radiomic features like GLCM and GLDM, and clinical variables like BMD T-score, are crucial in predicting hip fractures. This level of interpretability allows clinicians to understand which factors are most predictive of fracture risk, improving trust in the model's outcomes and facilitating informed decision-making.

Moreover, the proposed method is not limited to hip fracture risk assessment but can also be applied to other diagnostic tasks using medical images where the relationships between RoIs play a critical role. The versatility of the graph-based approach allows it to be extended to other conditions where structural relationships within medical images, such as in the analysis of bone density or tumor growth patterns, are essential.

**Acknowledgement**

This research has been conducted using the UK Biobank Resource under application number [61915]. It was in part supported by grants from the National Institutes of Health, USA (U19AG055373, 1R15HL172198, and 1R15HL173852) and American Heart Association (#25AIREA1377168).

**Reference**

[1] Tian C, Shi L, Wang J, Zhou J, Rui C, Yin Y, Du W, Chang S, Rui Y. Global, regional, and national burdens of hip fractures in elderly individuals from 1990 to 2021 and predictions up to 2050: A systematic analysis of the Global Burden of Disease Study 2021. Arch Gerontol Geriatr. 2025 Jun 1;133:105832.

[2] Amarilla-Donoso FJ, López-Espuela F, Roncero-Martín R, Leal-Hernandez O, Puerto-Parejo LM, Aliaga-Vera I, Toribio-Felipe R, Lavado-García JM. Quality of life in elderly people after a hip fracture: a prospective study. Health Qual Life Outcomes. 2020 Mar 14;18:71. PMCID: PMC7071575

[3] Aldieri A, Paggiosi M, Eastell R, Bignardi C, Audenino AL, Bhattacharya P, Terzini M. DXA-based statistical models of shape and intensity outperform aBMD hip fracture prediction: A retrospective study. Bone. 2024 May 1;182:117051.

[4] Kwon S, Yoo J, Yoon Y, Lee M, Hwang J. Clinical significance of trabecular bone score of DXA in hip fracture patients-comparative study between trochanteric fractures and neck fractures. BMC Musculoskelet Disord. 2024 Nov 13;25(1):908.

[5] Lex JR, Di Michele J, Koucheki R, Pincus D, Whyne C, Ravi B. Artificial Intelligence for Hip Fracture Detection and Outcome Prediction. JAMA Netw Open. 2023 Mar 17;6(3):e233391. PMCID: PMC10024206

[6] Krugh M, Langaker MD. Dual-Energy X-Ray Absorptiometry. StatPearls [Internet]. Treasure Island (FL): StatPearls Publishing; 2025 [cited 2025 Mar 28]. Available from: http://www.ncbi.nlm.nih.gov/books/NBK519042/ PMID: 30085584

[7] Classification of Fracture Risk in Fallers Using Dual-Energy X-Ray Absorptiometry (DXA) Images and Deep Learning-Based Feature Extraction - PMC [Internet]. [cited 2025 Mar 28]. Available from: https://pmc.ncbi.nlm.nih.gov/articles/PMC10731096/

[8] Terzini M, Aldieri A, Rinaudo L, Osella G, Audenino AL, Bignardi C. Improving the Hip Fracture Risk Prediction Through 2D Finite Element Models From DXA Images: Validation Against 3D Models. Front Bioeng Biotechnol. 2019;7:220. PMCID: PMC6746936

[9] Hsieh C-I, Zheng K, Lin C, Mei L, Lu L, Li W, Chen F-P, Wang Y, Zhou X, Wang F, Xie G, Xiao J, Miao S, Kuo C-F. Automated bone mineral density prediction and fracture risk assessment using plain radiographs via deep learning. Nat Commun. 2021 Sep 16;12:5472. PMCID: PMC8446034

[10] Shaik A, Larsen K, Lane NE, Zhao C, Su K-J, Keyak JH, Tian Q, Sha Q, Shen H, Deng H-W, Zhou W. A staged approach using machine learning and uncertainty quantification to predict the risk of hip fracture. Bone Rep. 2024 Sep;22:101805.

[11] Yan J, Yin X-C, Lin W, Deng C, Zha H, Yang X. A Short Survey of Recent Advances in Graph Matching. Proc 2016 ACM Int Conf Multimed Retr [Internet]. New York New York USA: ACM; 2016 [cited 2022 Jun 26]. p. 167–174. Available from: https://dl.acm.org/doi/10.1145/2911996.2912035

[12] Ma G, Ahmed NK, Willke TL, Yu PS. Deep graph similarity learning: a survey. Data Min Knowl Discov. 2021 May;35(3):688–725.

[13] Gu Y, Yang X, Tian L, Yang H, Lv J, Yang C, Wang J, Xi J, Kong G, Zhang W. Structure-aware siamese graph neural networks for encounter-level patient similarity learning. J Biomed Inform. 2022 Mar;127:104027.

[14] Zhao C, Esposito M, Xu Z, Zhou W. HAGMN-UQ: Hyper Association Graph Matching Network with Uncertainty Quantification for Coronary Artery Semantic Labeling. Med Image Anal. 2024 Oct;103374.

[15] Russell BC, Torralba A, Murphy KP, Freeman WT. LabelMe: a database and web-based tool for image annotation. Int J Comput Vis. Springer; 2008;77:157–173.

[16] Zhao C, Xu Z, Jiang J, Esposito M, Pienta D, Hung G-U, Zhou W. AGMN: Association graph-based graph matching network for coronary artery semantic labeling on invasive coronary angiograms. Pattern Recognit. 2023 Nov;143:109789.

[17] Zhao C, Xu Y, He Z, Tang J, Zhang Y, Han J, Shi Y, Zhou W. Lung segmentation and automatic detection of COVID-19 using radiomic features from chest CT images. Pattern Recognit. Elsevier; 2021;119:108071.

[18] Adams RP, Zemel RS. Ranking via Sinkhorn Propagation [Internet]. arXiv; 2011 [cited 2024 Jul 12]. Available from: http://arxiv.org/abs/1106.1925

[19] Cruz RS, Fernando B, Cherian A, Gould S. Visual Permutation Learning. IEEE Trans Pattern Anal Mach Intell. 2019 Dec 1;41(12):3100–3114.

[20] Hou Y, Hu B, Zhao WX, Zhang Z, Zhou J, Wen J-R, editors. Neural Graph Matching for Pre-training Graph Neural Networks. Proc 2022 SIAM Int Conf Data Min SDM [Internet]. Philadelphia, PA: Society for Industrial and Applied Mathematics; 2022 [cited 2024 Jul 11]. Available from: https://epubs.siam.org/doi/book/10.1137/1.9781611977172

[21] Wang R, Yan J, Yang X. Combinatorial Learning of Robust Deep Graph Matching: an Embedding based Approach. IEEE Trans Pattern Anal Mach Intell. 2020;1–1.

[22] Bottou L. Large-Scale Machine Learning with Stochastic Gradient Descent. In: Lechevallier Y, Saporta G, editors. Proc COMPSTAT2010 [Internet]. Heidelberg: Physica-Verlag HD; 2010 [cited 2025 Mar 16]. p. 177–186. Available from: http://link.springer.com/10.1007/978-3-7908-2604-3_16

**Supplementary materials**

**Table S1.** Comparison of graph generation methods in ICGM-Frax for hip fracture risk assessment.

| Graph generation method | ACC | F1 | SN | SP |
| --- | --- | --- | --- | --- |
| kNN | 0.9970± 0.0068 | 0.9932± 0.0162 | 0.9869±0.0310 | 1.0000± 0.0000 |
| Delaunay | 0.9889± 0.0138 | 0.9565± 0.0565 | 0.9214±0.0979 | 1.0000± 0.0000 |
| Distance | 0.9970± 0.0068 | 0.9886± 0.0258 | 0.9786±0.0482 | 1.0000± 0.0000 |